\documentclass[preprint,12pt,times]{article}

\usepackage{amssymb}
\usepackage{amsmath,graphicx}
\usepackage{graphicx}
\usepackage{multirow}
\usepackage{flushend}
\usepackage[table,xcdraw]{xcolor}
\usepackage{comment}
\usepackage{xcolor}
\usepackage{xspace}
\usepackage{algorithm}
\usepackage[noend]{algpseudocode} %
\usepackage{listings}
\usepackage{subfigure}
\usepackage{authblk}
\usepackage{url}

\title{Bridging Human Concepts and Computer Vision for Explainable Face Verification}

\author[1,2]{Miriam Doh~\footnote{work supported by the ARIAC project (No. 2010235),
funded by the Service Public de Wallonie (SPW Recherche).}}

\affil[1]{ISIA lab - Université de Mons (UMONS), Bd Dolez 31, 7000,
   Mons, Belgium}
\affil[2]{IRIDIA lab - Université Libre de Bruxelles (ULB),
  Av. Adolphe Buyl 87,1060, Ixelles, Belgium}

\author[3,4]{Caroline Mazini Rodrigues}

\affil[3]{LRE - Laboratoire de Recherche de l'EPITA, 14-16 Rue Voltaire, 94270, Kremlin-Bicêtre, France}
\affil[4]{LIGM - Laboratoire d'Informatique Gaspard-Monge, Université Gustave-Eiffel, 77454, Marne-la-Vallée, France}

\author[3]{Nicolas Boutry}
\author[4]{Laurent Najman}
\author[1]{Matei Mancas}

\author[2]{Hugues Bersini}

\date{} 
\providecommand{\keywords}[1]{\textbf{\textit{Keywords---}} #1}

\begin{document}

\maketitle

\begin{abstract}
With Artificial Intelligence (AI) influencing the decision-making process of sensitive applications such as \textit{Face Verification}, it is fundamental to ensure the transparency, fairness, and accountability of decisions. Although Explainable Artificial Intelligence (XAI) techniques exist to clarify AI decisions, it is equally important to provide interpretability of these decisions to humans. In this paper, we present an approach to combine computer and human vision to increase the explanation's interpretability of a face verification algorithm. In particular, we are inspired by the human perceptual process to understand how machines perceive face's human-semantic areas during face comparison tasks. We use Mediapipe, which provides a segmentation technique that identifies distinct human-semantic facial regions, enabling the machine's perception analysis. Additionally, we adapted two model-agnostic algorithms to provide human-interpretable insights into the decision-making processes.
\end{abstract}


\section{Introduction}
Face verification \cite{1alfarsi2019techniques} aims to confirm an individual's identity based on facial features, with applications in law enforcement \cite{20lynch2020face}, border control \cite{10del2016automated}, or smartphone security \cite{26robertson2015face}. As AI becomes prevalent in decision-making \cite{37zhang2022ai}, ensuring model fairness, accountability, confidentiality, and transparency is crucial \cite{22olteanu2021facts}. However, complex ML models are often seen as 'black boxes' \cite{7davide2016can}. Explainable AI (XAI) \cite{guidotti2018survey} addresses this challenge by enhancing AI interpretability to make AI systems transparent and understandable to humans, thereby increasing trust in their decisions.

Saliency maps have become the most popular XAI solution in computer vision, offering insights into the critical features considered in the decision-making process. However, in face verification, decisions often rely on adjustable thresholds based on the specific application rather than understandable semantic classes. This raises questions about the adequacy of identifying the most important features in an image as the only explanation \cite{kim2023help}.
Taking inspiration from the human perceptual process, we propose a model-agnostic approach capable of determining how the machine perceives similar semantic areas of the face when comparing two faces. Our primary objective is to translate the XAI solution into human decision-making meaningfully. However, incorporating human-based semantics in the models' explanation process can also introduce human bias to these same explanations. To increase human interpretability, we must also assure the \textit{Faithfulness} of explanations to the model's reasoning. \textit{Faithfulness} refers to whether a feature, considered important for the model, changes the model's decision~\cite{bommer2023corr}. %

For face verification, the model extracts features for each face that will be compared. Modifications in the features will also impact how similar are the two faces. Therefore, it is essential to understand how face parts, such as an eye, would impact the final features. 

To translate the model's knowledge to human knowledge as smoothly as possible, we first perform the segmentation of face parts based on human semantics. By considering the impact of those face parts on a set of face images, we can have a global view of the model's knowledge. %
Following the features' extraction (through the model), we verify if two people are the same by comparing their facial features. To understand the contribution of the chosen \textit{concepts} to the relation between two compared faces, we introduce an algorithm grounded in the perturbation of facial regions linked to the extracted concepts, mirroring the human perceptual process of face recognition. It encompasses evaluating corresponding semantic areas along a spectrum of similarities, providing interpretation and contextualisation. %

We structured this paper as follows: in Section~\ref{sec:sota}, we present state-of-the-art methods for explaining the face verification task; in Section~\ref{sec:method}, we describe our framework, including the model concept's extraction and the perturbation methods for face comparison; in Section~\ref{sec:experiments} we include the experimental results and limitations; in Section~\ref{sec:conclusion} we conclude the work.

\section{Related work}
\label{sec:sota}

Saliency maps, such as CAMs \cite{38zhou2016learning,28selvaraju2017grad} and RISE  \cite{23petsiuk2018rise}, are crucial for interpreting deep-learning models, revealing their inner workings. However, their primary development has centered on object recognition, leaving the field of face analysis relatively unexplored.

Despite its critical applications, research in face analysis has been limited. Works by \cite{36yin2019towards,21mery2022black,19knoche2023explainable} mainly focus on individual pixel or low-level feature significance, which can be challenging for human analysts and may not align with intuition.
Conversely, LIME \cite{25ribeiro2016should} employs superpixels within the image, providing a user-friendly, concept-driven explanation. However, this technique relies on a new model approximating the original, potentially obscuring the actual reasons for the original model's behavior \cite{Ribera2019CanWD}.

Alternative approaches, such as TCAV~\cite{kim:icml:2018} and knowledge graphss~\cite{zhang:aaai:2018}, prioritize low-level importance from pixels and aim to represent the model's knowledge through \textit{concepts}. TCAV employs semantic concepts defined by users or discovered through image segment activations (with method ACE~\cite{ghorbani:2019:neurips}), while knowledge graphs identify repeating patterns across network layers. Additionally, Tan et al.~\cite{tan:2022:neuralNets} introduced the Locality Guided Neural Network (LGNN), designed to induce filter topology that enhances the visualization of concepts.

Inspired by these methods, our approach combines human and model perspectives to identify essential concepts for face verification. We acknowledge that relying solely on human concepts can introduce bias while relying solely on the model can complicate interpretation.

\section{Proposed Method}
\label{sec:method}
\begin{figure}[ht]
    \centering
    \includegraphics[width=\textwidth ]{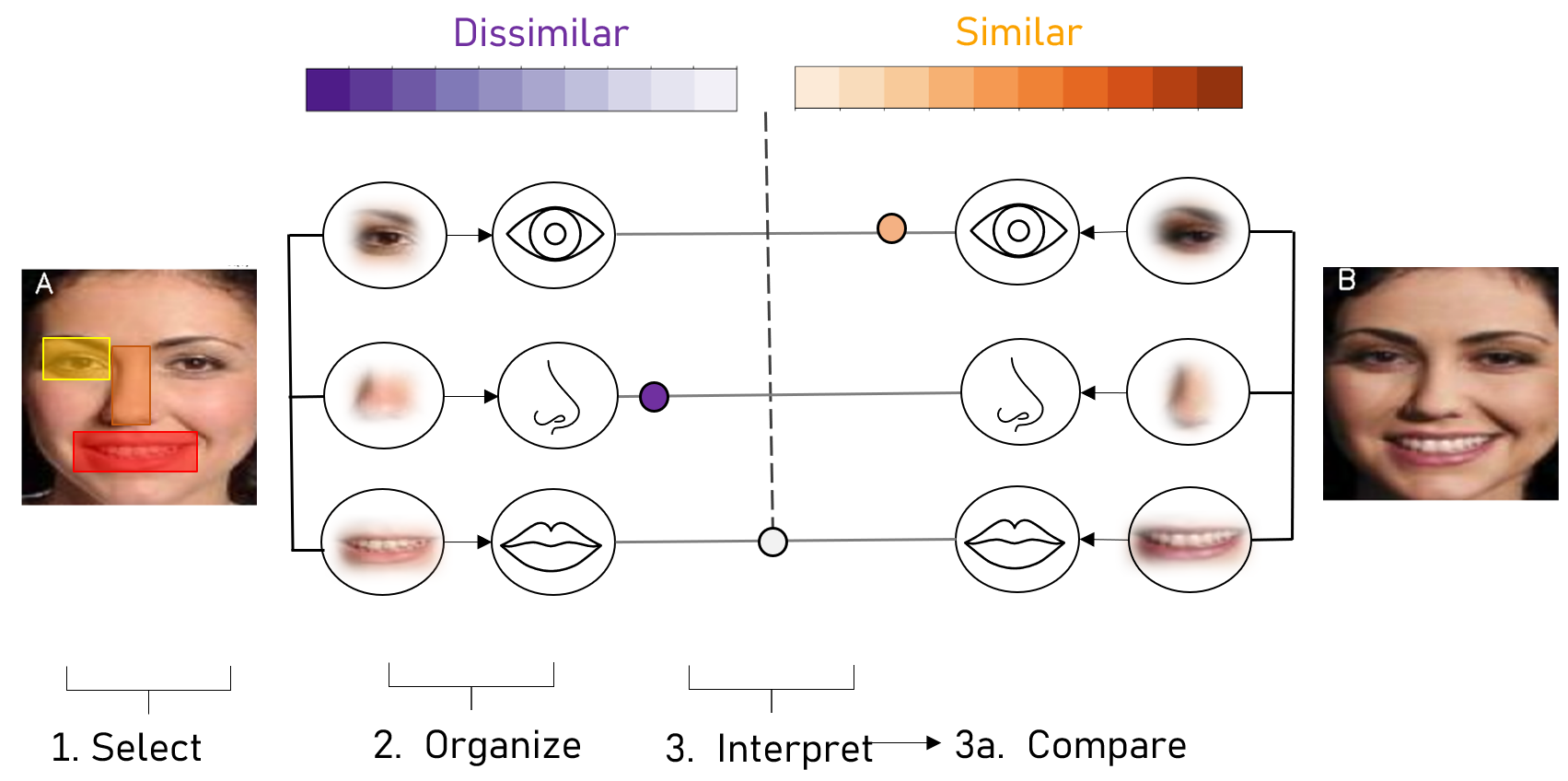}
    \caption{Face verification adaptation of XAI Perceptual processing framework proposed by \cite{34zhang2022towards} and inspired by how humans process stimuli (select, organize, interpret and compare)}
    \label{fig: framework}
\end{figure}
To help humans understand how AI systems make decisions, it is essential to present the information in a way that aligns with human cognitive processes. Cognitive psychology provides valuable insights into how people perceive and process information when identifying faces. Taking inspiration from the flowchart proposed by \cite{34zhang2022towards}, we aim to apply a similar method to face verification (see Figure \ref{fig: framework}).
The human perceptual process consists of three key phases: selection, organization, and interpretation \cite{alma9924474754502466}. Cognitive psychology has shown that when recognizing faces, our attention is particularly drawn to particular facial areas, such as the eyes and nose \cite{3matthews1978discrimination,4davies1977cue,5iskra2016eye}. 
Subsequently, in the perceptual process, these facial stimuli are organized into meaningful concepts, 
adding semantics to the process. Our brains compare these higher-level concepts to assess the similarity between items, facilitating face categorization. This comparative analysis may involve matching a face to a remembered image or with another face in front of us. %
In this context, we question the adequacy of salience maps used in computer vision as an explanation and their alignment with our human reasoning processes.

\begin{figure}[ht]
    \centering
    \includegraphics[width=\textwidth]{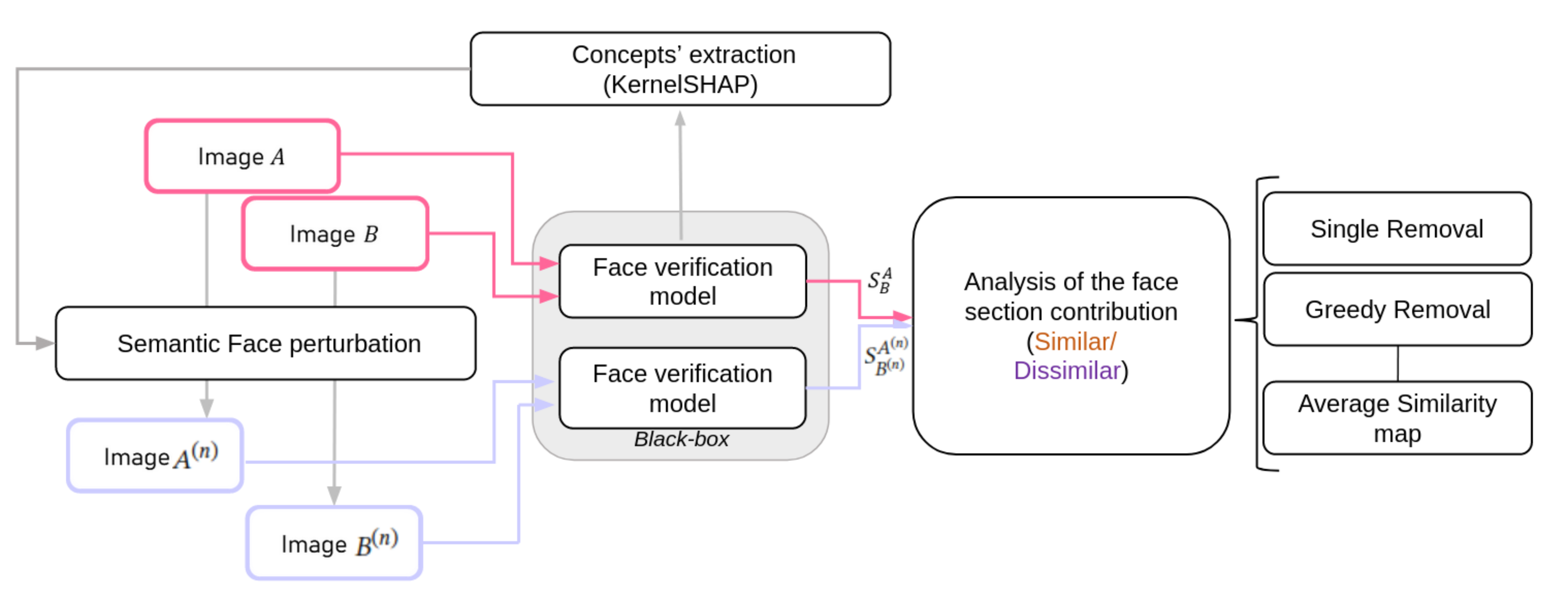}
    \caption{Proposed flowchart. We extract concepts from the face verification model (using KernelSHAP) and input them into a Semantic Face perturbation phase. In this phase, the two images' perturbation is made in the same regions to evaluate similarities and dissimilarities. We propose three algorithms for the perturbations: Single removal, greedy removal, and average similarity map.}
    \label{fig: flowchart}
\end{figure}
Based on cognitive psychology, we have developed a general flowchart shown in Figure \ref{fig: flowchart}. %
Generally, face verification systems rely primarily on a matching score between two face images $A$ and $B$. This score, $S^{A}_B$, is computed using cosine similarity, which compares the feature vectors $\mathbf{f_A}$ and $\mathbf{f_B}$ extracted from each image as follows:
\begin{equation}
 S^{A}_B = \frac{\mathbf{f_A} \cdot \mathbf{f_B}}{||\mathbf{f_A}||\,||\mathbf{f_B}||}
\end{equation}

The resulting score ranges from 0 to 1, with a score of 1 indicating identical images ($A = B$).
As our approach is model-agnostic, we aim to explain the algorithm by perturbing the inputs to study the system's decision behaviour concerning the input-output relationship. Inspired by the work of \cite{19knoche2023explainable}, our desired output is a similarity map indicating which face areas are considered similar or dissimilar for both images, using an AI model as a feature extractor. To achieve this, we perform semantic perturbation on images $A$ and $B$, resulting in new images denoted as $A^{(n)}$ and $B^{(n)}$ where the $n$ section is removed in both images. We obtain a new $S^{A^{(n)}}_{B^{(n)}}$ score from these perturbed images. By fairly masking the images, we can assess if the system perceives semantic areas, such as the eyes, as similar or dissimilar. Considering $\Delta_S$ the difference between original and new scores represented by Equation~\ref{eq:delta}, if the $S^{A^{(n)}}_{B^{(n)}}$ decreases compared to $S^{A}_{B}$, it suggests that the removed parts positively contribute to the similarity ($\Delta_S \geq 0$). Conversely, its increase indicates a negative contribution ($\Delta_S < 0$).

\begin{equation}
  \Delta_S =S^{A}_{B} - S^{A^{(n)}}_{B^{(n)}}  
  \label{eq:delta}
\end{equation}

\begin{figure}[ht]
    \centering
    \includegraphics[width=\textwidth ]{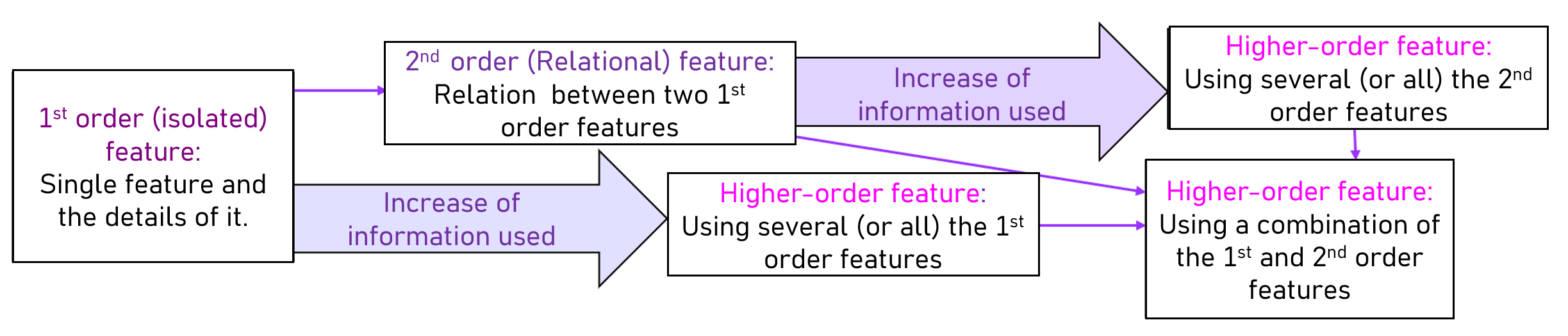}
    \caption{An interpretation of a relational/configural model of face perception.}
    \label{fig: processing}
\end{figure}

Compared to \cite{19knoche2023explainable}, our objective is to incorporate semantic masking in the perturbation process to increase interpretability by providing not only a map but also a chart related to the semantic areas. We apply two types of perturbation algorithms inspired by \cite{21mery2022black}, allowing us to study the face section area's single or collaborative contributions and then incorporate this information within an average similarity map. This single/collaborative approach aligns with the notion that humans perceive and interpret faces in a relational/configurational way \cite{rhodes1993configural} (see Figure \ref{fig: processing}). First-order features concern individual components that can be processed independently (e.g., eyes, nose). Second-order features involve information acquired when simultaneously processing two or more parts together (e.g., spacing between eyes). Furthermore, higher-order features emerge from combinations of multiple first-order and/or second-order features. In our case, the single removal procedure models the information associated with first-order or single features, and the greedy removal procedure addresses the second-order features, wherein multiple parts are processed collectively.
\subsection{Semantic Extraction}
\begin{figure}[ht]
    \centering
    \includegraphics[width=0.5\textwidth ]{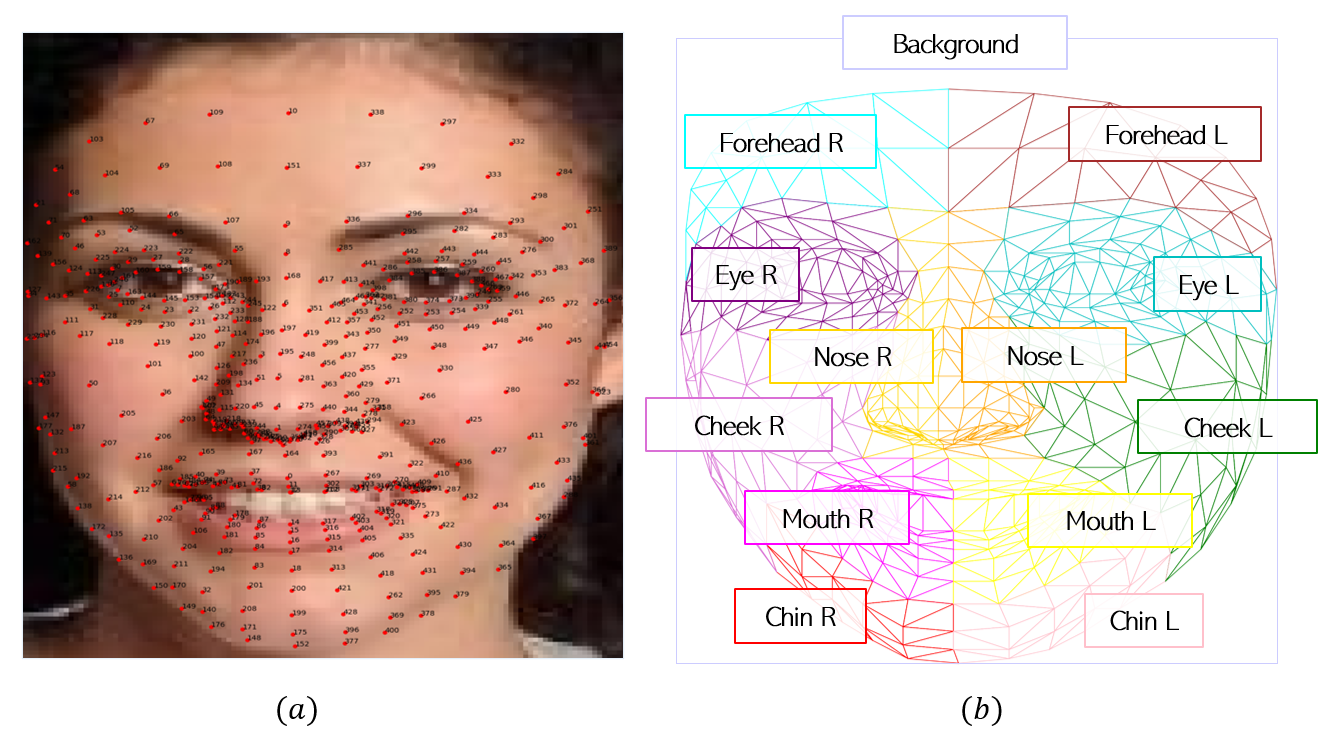}
    \caption{In the image (a) Mediapipe landmarks are plotted on the sample image. In the image (b), the 13 semantic sections are defined through the landmarks}
    \label{fig: mp}
\end{figure}
To incorporate semantics, we employ Mediapipe Face masks, a versatile open-source framework by Google, widely recognized for its face detection and landmark estimation capabilities. By extracting landmarks from Mediapipe, we defined 13 polygons corresponding to distinct semantic areas of the face (see Figure \ref{fig: mp}.a). The landmark estimation provided by Mediapipe is limited to specific facial regions, and hair or ears were not included in the earlier facial subdivisions. Nevertheless, this decision is consistent with previous research \cite{15Karczmarek}, which demonstrated that some areas of the face are more influential than others. For example, removing the ears has less impact on the final score than the eye area. Hence, we assumed these areas were not primarily influential and did not include them in our face classes. Additionally, face verification algorithms typically apply a preprocessing step for extracting the face area. Therefore, we reduce the area outside the face by applying MCTNN \cite{16li2020face}, a deep learning-based face detection algorithm.
Overall, our subdivision of the face detected 13 distinct semantic classes, including the background (figure \ref{fig: mp}.b).
With this approach, the semantic areas vary in size, resulting in larger maps having a more significant influence on the score than smaller ones.
To mitigate this undesired effect, we introduce a weight, denoted as ${w}_{A,n}$  related to the section $n \in [1,m]$ with $m=13$. The ${w}_{A,p}$ is defined as the rapport between the total area of the image $A$ ($\mathit{Area}_A$) and the area of the mask $M^{(A,n)}$, $\mathit{Area}_{M^{(A,n)}}$, indicating region $p$ (white pixels in the mask). This weight serves to counterbalance the differences in magnitude. Moreover, due to the precise face positioning achieved by Mediapipe, the masks obtained on images $A$ and $B$ may only partially match. This discrepancy arises because the depicted faces may not have the same position and expressiveness. For this reason, we define two weight ${w}_{A,n} = W(A,M^{(A,n)}) = \frac{\mathit{Area}_A}{\mathit{Area}_{M^{(A,n)}}}$ associated to $A$ and ${w}_{B,n} = W(B,M^{(B,n)}) = \frac{\mathit{Area}_B}{\mathit{Area}_{M^{(B,n)}}}$ to $B$.

\begin{align}
\label{eqn:W}
\begin{split}
\widehat{W}_{(A,B)_{n}}=\frac{w_{A,n}. w_{B,n}}{\sum_{i=1}^{m} w_{A,i} . w_{B,i}}
\end{split}
\end{align}
\begin{align}
\label{eqn:C}
\begin{split}
C_{n} =\Delta_S \cdot \widehat{W}_{(A,B)_{n}} 
\end{split}
\end{align}
In this manner, the contribution of the mask, defined as $C_n$, is weighted by $w_{A,n}$ and $w_{B,n}$, representing the relative weights associated with $M^{(A,n)}$ and $M^{(B,n)}$ masks, respectively.

\subsubsection{Concepts Extraction}

Using Mediapipe for face part extraction provides a human-based semantic segmentation, yet it may not align with how models perceive faces. To bridge this gap we introduce a model's concept extraction process. This involves filtering machine-important parts based on human semantics. 
For evaluating the importance of facial parts, we employ KernelSHAP~\cite{lundberg2017nips}, which combines LIME~\cite{25ribeiro2016should}'s interpretable components with Shapley values~\cite{castro2009cor} from game theory which look for each feature contribution to the final result. We extract model importance scores for each of semantic parts. %
In face final representations with 512 features, for example, we will have 512 importance scores per human-semantics part. 
In the process of face verification, every feature change, negative or positive, is significant to determine faces' similarity, with emphasis on the magnitude of the change, instead of on the signal. If one feature of a human-semantic part obtained a negative Shap value, the lack of this part reduced the feature value, and vice versa. Therefore, negative and positive Shap values are equally important in our context. For this reason, sum the absolute Shap values throughout all the representation features to obtain a single importance value per part.

\begin{figure}[!ht]
\centering
\begin{tabular}{cc}
\includegraphics[width=6.5cm]{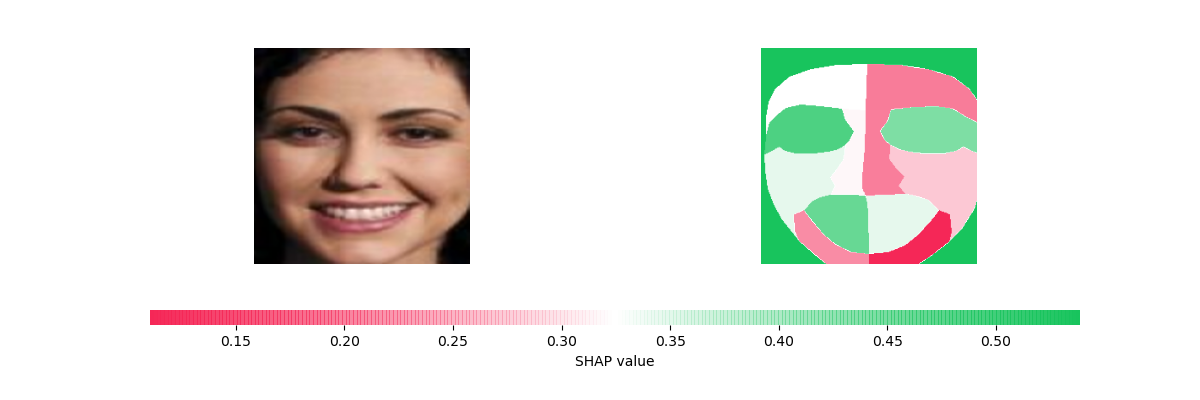}
&
\includegraphics[width=6.5cm]{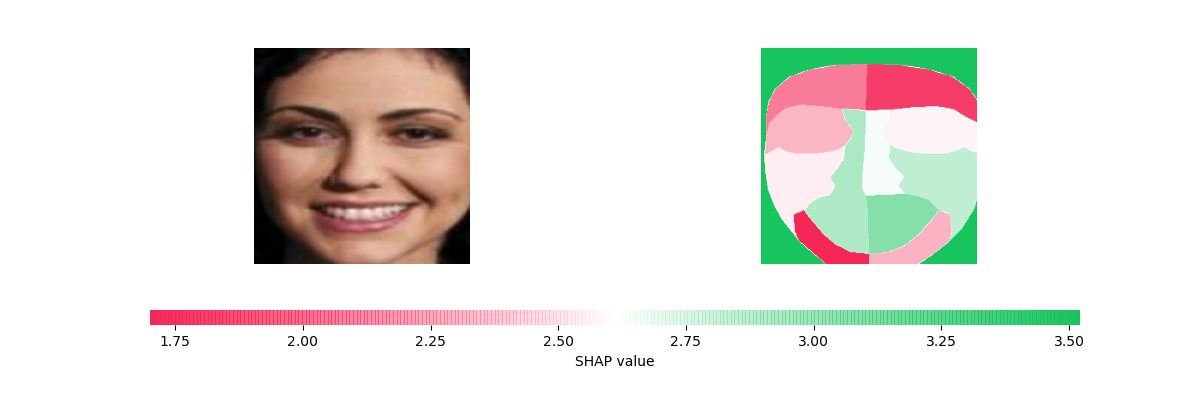}\\
(a) & (b) \\
\includegraphics[width=6.5cm]{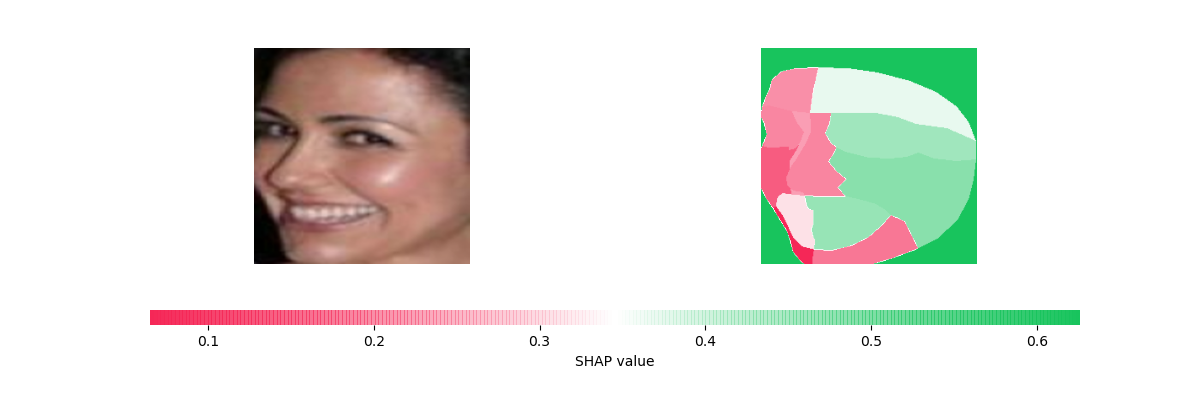}
& 
\includegraphics[width=6.5cm]{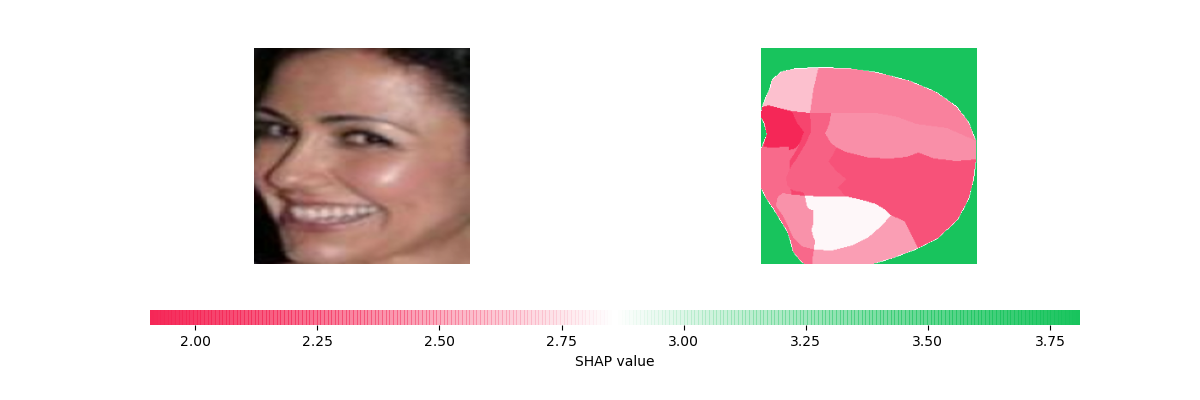}\\
(c)  & (d) 
\end{tabular}
\caption{Examples of two images' human-semantics part importance scores using KernelSHAP~\cite{lundberg2017nips}. We analyse two models: CasiaNet~\cite{yi2014learning} in (a) and (c), and VGGfaces2~\cite{massoli2020ivc} in (b) and (d). Green parts are more important according to Shap scores. There are differences between important parts for different images, especially for VGGfaces2. That is why we aggregate ranked importance over 200 images.}
\label{fig:concepts_extraction}
\end{figure}

Ultimately, we will have one importance value per semantic part (see Figure~\ref{fig:concepts_extraction}). However, this remains a local importance, i.e., an importance score according to a single image dataset. To increase globalism in the concepts' extraction, we need to include information from multiple images.
Our solution is to combine the importance levels from a set of images by a ranking combination strategy. Each image obtains 13 importance scores (one per human-semantics part) that we can order. More significant scores are at the top of a ranking, as they were considered more important for the model. We use 200 images from  CelebA~\cite{zhang2018arxiv} dataset to obtain 200 rankings. From these rankings' combinations, by a vote-based technique using BORDA count~\cite{borda1781}, we obtain a final ranking with more globally important concepts at the top.

The experiments will focus on the model's top eight concepts determined by this process.

\subsubsection{Proposed similarity maps}
The algorithm used to generate the similarity maps draws inspiration from the work of \cite{21mery2022black}, where six algorithms were presented to create saliency maps. Specifically, we will employ the single removal approach (S0) and the greedy removal approach (S1), with the possibility of creating an average map of these two approaches (S\textsubscript{AVG}). Our approach incorporates significant changes compared to previous research.
First and foremost, we utilize semantically meaningful masks to perturb the images, diverging from conventional circular or square masks with a fixed shape. Moreover, since our objective is to generate community similarity maps between the two images, both images undergo perturbation, contrary to previous approaches that typically perturb only one of the images, thus aligning more closely with the strategy proposed by \cite{19knoche2023explainable}.
\subsubsection{Single Removal - S0}
\label{subsec:SR}
\begin{figure}[ht]
    \centering
    \includegraphics[width=\textwidth ]{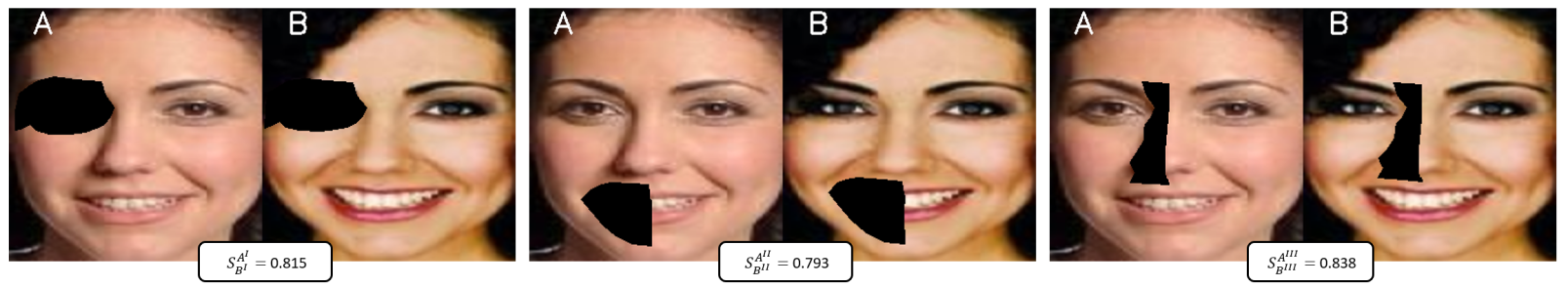}
    \caption{Samples of single removal where $S^{A^{(n)}}_{B^{(n)}}$ is the cosine similarity between image $A$ and image $B$ with the $n$ semantic part removed. }
    \label{fig: single_removal}
\end{figure}

We define the two perturbed images as the pixel-wise multiplication of the images and the relative semantic section mask of the same size with values between 0 and 1.
\begin{align}
\label{eqn:AB_perturbation}
\begin{split}
A'= A \cdot M^{(A,n)}
\text{\ \ and \ \ }
B'= B \cdot M^{(B,n)}
\end{split}
\end{align}
The single removal operation is computed for all the semantic areas. For each mask, the value of the contribution map H0 is initialized with the $C_{n}$ contribution associated with the mask:
\begin{align}
\label{eqn:H0}
\begin{split}
H0_{(A,n)} = C_n \cdot M^{(A,n)} %
\end{split}
\end{align}

The similarity map is defined as the sum of the negative and the positive contributions normalized by Equation~\ref{eqn:S0} for all $n \in [1,m]$, to obtain $S0_A$.

\begin{align}
\label{eqn:S0}
\begin{split}
 H0_{(A,n)}^{\pm} &=
\begin{cases}
 \frac{H0_{(A,n)}}{\sum_{\substack{ H0_{(A,m)} \geq 0}} \lvert H0_{(A,m)}\rvert} \quad \text{if } H0_{(A,n)}\geq 0 \\
  \frac{H0_{(A,n)}}{\sum_{\substack{H0_{(A,m)}< 0}} \lvert H0_{(A,m)}\rvert} \quad \text{otherwise.}\\%\text{if } H0_{(A,n)} < 0 
 \end{cases}\\
S0_A &= S0_A + (H0_{(A,n)}^{+} + H0_{(A,n)}^{-}) \cdot M^{(A,n)}
\end{split}
\end{align}

We use the same Equations~\ref{eqn:H0}~and~\ref{eqn:S0} to obtain $H0_{(B,n)}, H0_{(B,n)}^+, H0_{(B,n)}^-$ and $S0_B$. This means negative contributions are seen as dissimilar areas in the face image, while positive ones are similar. The algorithm \ref{alg:the_alg} gives us the similarity maps $S0_A$ and $S0_B$ as a result of single removal.

\begin{algorithm}
\caption{Calculate H0 and H1}\label{alg:the_alg}
\tiny
\begin{algorithmic}[1] %
\State \textbf{Input:}
\State $A$  -Face image A
\State $B$ -Face image B
\State $S^{A}_{B}$ -Initial Score
\State $\theta$ -Minimal increment allowed
\State $t_{max}$ -Max number of iteration
\Statex \hrulefill
\State N, M $\leftarrow$ size($A$) \Comment{height and width of face images}
\State $A_0 \leftarrow A$  \Comment{initialization of the image}
\State $B_0 \leftarrow B$  \Comment{initialization of the image}
\State H0\textsubscript{$A$} $\leftarrow$ zeros(N, M) \Comment{initialization of the maps}
\State H0\textsubscript{$B$} $\leftarrow$ zeros(N, M) \Comment{initialization of the maps}
\State H1\textsuperscript{-}\textsubscript{$A$} $\leftarrow$ zeros(N, M) \Comment{initialization of the maps}
\State H1\textsuperscript{-}\textsubscript{$B$} $\leftarrow$ zeros(N, M) \Comment{initialization of the maps}
 
\State Best\textsubscript{M\textsubscript{$A$}} $\leftarrow$ zeros(N, M)\Comment{initialization of mask A}
\State Best\textsubscript{M\textsubscript{$B$}} $\leftarrow$ zeros(N, M)\Comment{initialization of mask B}
\State $t=0$ \Comment{initialization of iteration counter}
\State $s_{t-1} \leftarrow S^{A}_{B}$ \Comment{initial matching score}
\State $\Delta_s \leftarrow 1$ \Comment{initialization of difference of scores}
 
\While{$t$ < $t_{max}$ and $\Delta_s$ > $\theta$}
    \State $s_t \leftarrow 1$ 
    \State $t \leftarrow t+1$
 
    \For{$n$ in Face\textsubscript{Sections}}
        \State $M^{(A_t)}$ $\leftarrow$   $M^{(A_t,n)}$ +  Best\textsubscript{M\textsubscript{$A$}}
        \State $M^{(B_t)}$ $\leftarrow$   $M^{(B_t,n)}$ +  Best\textsubscript{M\textsubscript{$B$}}
        \State $A'$= $A$\textsubscript{$t-1$} $\cdot$ $M^{(A_t)}$
        \State $B'$ = $B$\textsubscript{$t-1$} $\cdot$ $M^{(B_t)}$
        \State $s'\leftarrow S^{A^{'}}_{B^{'}}$
 
        \If{$s' < s_t$}
            \State $s_t = s'$
            \State Best\textsubscript{M\textsubscript{$A$}} $\leftarrow$ $M^{(A_t)}$
            \State Best\textsubscript{M\textsubscript{$B$}} $\leftarrow$ $M^{(B_t)}$
            \State Best\textsubscript{$w_A$} $\leftarrow$  W($A$,Best\textsubscript{M\textsubscript{$A$}})
            \State Best\textsubscript{$w_B$} $\leftarrow$  W($B$,Best\textsubscript{M\textsubscript{$B$}})
            \State $A$\textsubscript{t} $\leftarrow  A'$
            \State $B$\textsubscript{t} $\leftarrow  B'$
        \EndIf
    \EndFor
 
    \State $\Delta_{s_{t}}= s_{t-1} - s_{t}$
    \If{$t=0$}
        \For{$n$ in Face\textsubscript{Sections}}
            \State $C_{n} =  \Delta_{s_{t}} \cdot \widehat{W}_{{(A,B)}_{n}}   $
            \State $H0_A[M^{(A,n)}=1]$ $\leftarrow C_{n}$
            \State $H0_B[M^{(B,n)}=1]$ $\leftarrow C_{n}$
        \EndFor
    \EndIf
 
    \State $C_{\text{Best}} \leftarrow \Delta_{s_{t}} \cdot \widehat{W}_{{(A,B)}_{\text{Best}}}$
    \State H1\textsuperscript{-}\textsubscript{$A$}[Best\textsubscript{M\textsubscript{$A$}}=1] $\leftarrow C_{\text{Best}}$
    \State H1\textsuperscript{-}\textsubscript{$B$}[Best\textsubscript{M\textsubscript{$B$}}=1] $\leftarrow C_{\text{Best}}$
\EndWhile
\Statex \hrulefill
\State \textbf{Output:} H0\textsubscript{$A$},H1\textsuperscript{-}\textsubscript{$A$},H0\textsubscript{$B$},H1\textsuperscript{-}\textsubscript{$B$}
\end{algorithmic}
\end{algorithm}

\subsubsection{Greedy Removal-S1}
\label{subsec:GR}
\begin{figure}[ht]
    \centering
    \includegraphics[width=\textwidth ]{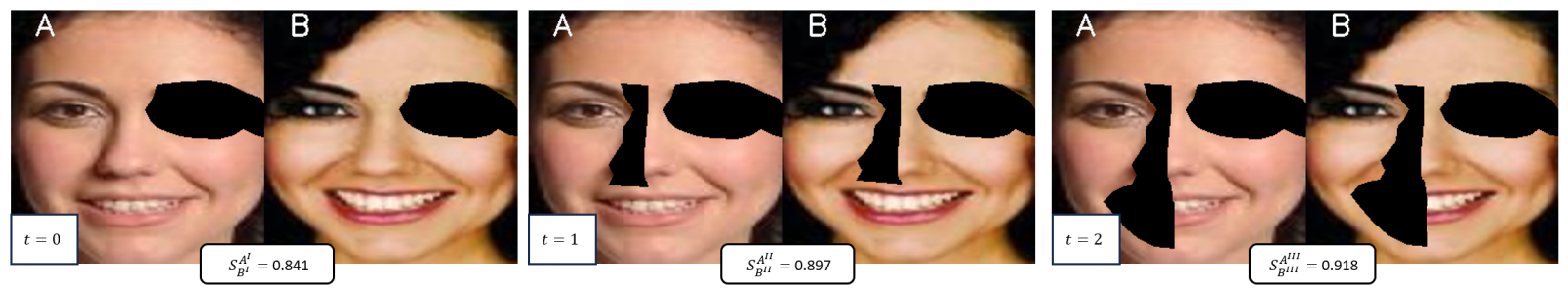}
    \caption{Greedy removal for image $A$ and image $B$ in $n$ steps ($t=n-1$), where $S^{A^{(n)}}_{B^{(n)}}$ is the cosine similarity between the two images and $n$ is the best part removed ($Best\textsubscript{M\textsubscript{$A$}}$ and $Best\textsubscript{M\textsubscript{$B$}}$) at $t$-step.}
    \label{fig: greedy_removal}
\end{figure}
The iterative approach of the greedy algorithm involves repeatedly performing a single removal procedure. In each iteration, the section of the face with the greatest impact is removed from images $A$ and $B$. In particular, the initial images are represented as $A_{0} = A$ and $B_{0} = B$, and at each iteration, $A_t$ and $B_t$ are obtained by removing the principal parts of $A_{t-1}$ and $B_{t-1}$, respectively. This means that at each iteration, the mask removed will be defined as the actual section mask sum with the previous best mask removed:
\begin{align}
\label{eqn:greedy_AB}
\begin{split}
M^{(A_t)} =  M^{(A_t,n)}  +  \text{Best\textsubscript{M\textsubscript{$A$}}}
\text{\ \ and \ \ }
M^{(B_t)} =  M^{(B_t,n)} +  \text{Best\textsubscript{M\textsubscript{$B$}}}
\end{split}
\end{align}
In greedy removal, calculating positive and negative contribution maps follows distinct procedures. We also use Equations~\ref{eqn:H0}~and~\ref{eqn:S0} to obtain $H1_{(.,n)},$ $ H1_{(.,n)}^+, H1_{(.,n)}^-,S1_A$ and $S1_B$. To be more concise, in algorithm \ref{alg:the_alg}, the presentation primarily focuses on calculating the negative contribution map H1\textsuperscript{-}. Indeed, in each iteration, the $s_t$ value is set to 1. Consequently, the removed areas correspond to those exhibiting negative contribution, as the condition $s'<s_t$ dictates. Conversely, H1\textsuperscript{+} is computed, setting $s_t$ value to 0 at each iteration with the condition $s'>s_t$.
In our example, the iteration stops when the maximum number of iterations is reached or when the score difference reaches a low enough point. In this case, that occurs at t = 7, where the score difference is only 0.009. After obtaining H1\textsuperscript{+} and H1\textsuperscript{-}, the similarity map S1 is obtained following the equation \ref{eqn:S0}.
\subsubsection{Average Similarity map S\textsubscript{AVG}}
In subsection \ref{subsec:SR} and \ref{subsec:GR}, the processes for determining similarity maps are outlined. Using single and greedy removal techniques makes it possible to assess the significance of each facial feature individually or in conjunction with others. Considering this, analyzing an average map can provide valuable comprehension of the significance of each facial feature.
Incorporating this information within an average similarity map of maps $S0$ (Single removal) and $S1$ (Greedy removal) which is called $S_{AVG}$ aligns with the notion that humans perceive and interpret faces in a relational/configurational manner \cite{rhodes1993configural} (see Figure \ref{fig: processing}). 

\section{Experimental results}
\label{sec:experiments}
\begin{figure}[ht]
    \centering
    \includegraphics[width=\textwidth ]{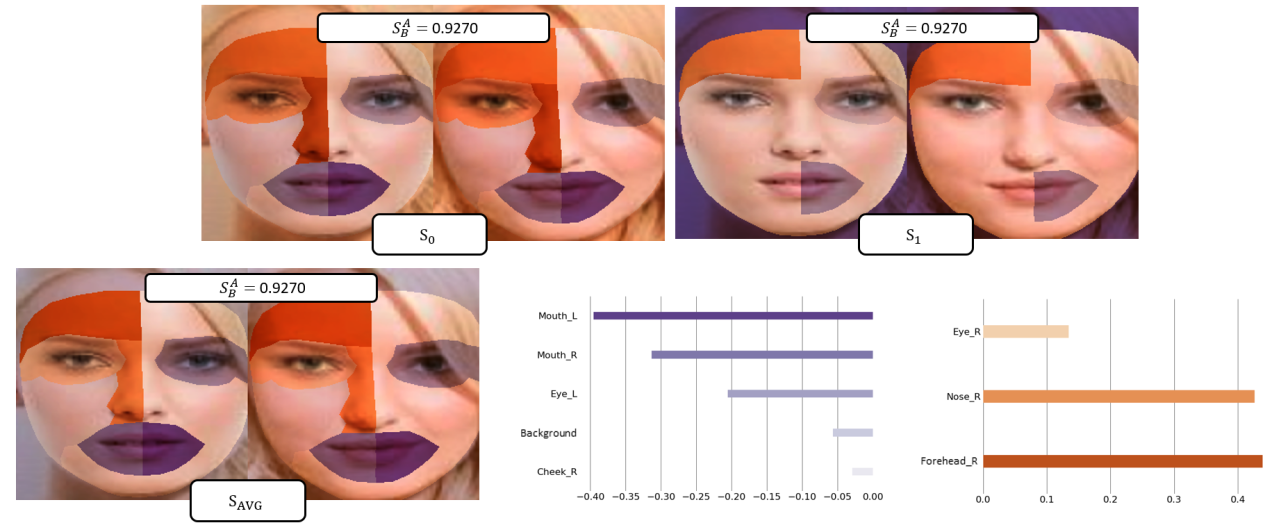}
    \caption{Similarity maps for each algorithm proposed in the case of VGGface2. Respectively S\textsubscript{0} is the output of the single removal, S\textsubscript{1} is the greedy removal ones, and S\textsubscript{AVG} is the average map generated from S\textsubscript{0}, and S\textsubscript{1}. The plot chart considers the contribution values ($C_n$) for each section in the perturbation.}
    \label{fig: maps}
\end{figure}
This section shows the experimental results for a selected number of samples extracted by the CelebA dataset \cite{zhang2018arxiv} and tested for the FaceNet \cite{schroff2015facenet} model trained on Casia-WebFace \cite{yi2014learning} and VGGfaces2 \cite{massoli2020ivc}. In Figure \ref{fig: maps}, we show the output generated by the proposed method. It comprises three maps: the initial single removal map S\textsubscript{0}, the greedy removal map S\textsubscript{1}, and the ultimate average map S\textsubscript{AVG}. The visualization uses orange to represent semantic areas that are similar, while purple indicates differences in facial features.
After the concept extraction, a group of semantic areas is selected based on their importance. The table \ref{tab:tabella} displays the n-selected semantic areas ranked by their importance. %
In our study $n=8$, this number can be changed as needed.

\begin{table}[ht]
\centering
\caption{Concept extraction output of each model's top n semantic areas (n=8). Area names are abbreviated with the initial or the first two letters (i.e., E = eye). R and L are associated with the right and left sides.}
\begin{tabular}{|c|c|}
\hline
\cellcolor{gray!25} VGGFace2 & "B", "CHE\textsubscript{R}", "MO\textsubscript{L}", "E\textsubscript{R}", "MO\textsubscript{R}", "N\textsubscript{R}", "F\textsubscript{R}", "E\textsubscript{L}"\\
\hline
\cellcolor{gray!25} Casia Net & "B", "E\textsubscript{R}", "M\textsubscript{R}", "M\textsubscript{L}", "E\textsubscript{L}", "CHE\textsubscript{R}", "F\textsubscript{R}", "CHE\textsubscript{L}"\\
\hline
\end{tabular}
\label{tab:tabella}
\end{table}

\begin{figure}[ht]
    \centering
    \includegraphics[width=\textwidth ]{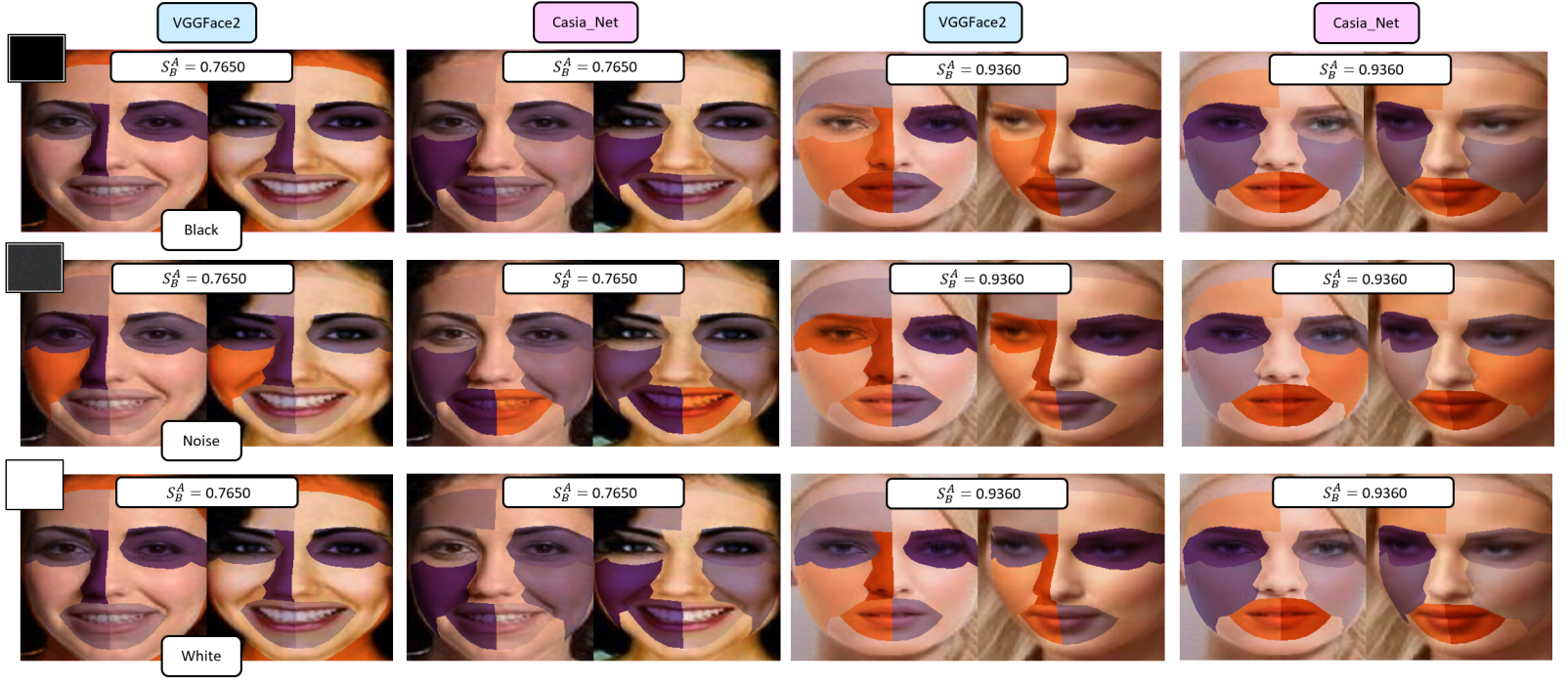}
    \caption{The S\textsubscript{AVG} map for pair examples from the CelebA dataset, generated using different patch coloring for the models VGGface2 and CasiaNet Orange hues denote similar facial regions, while purple highlights dissimilar ones}
    \label{fig: masking_sensitivity}
\end{figure}

\begin{figure}[ht]
    \centering
    \includegraphics[width=\textwidth ]{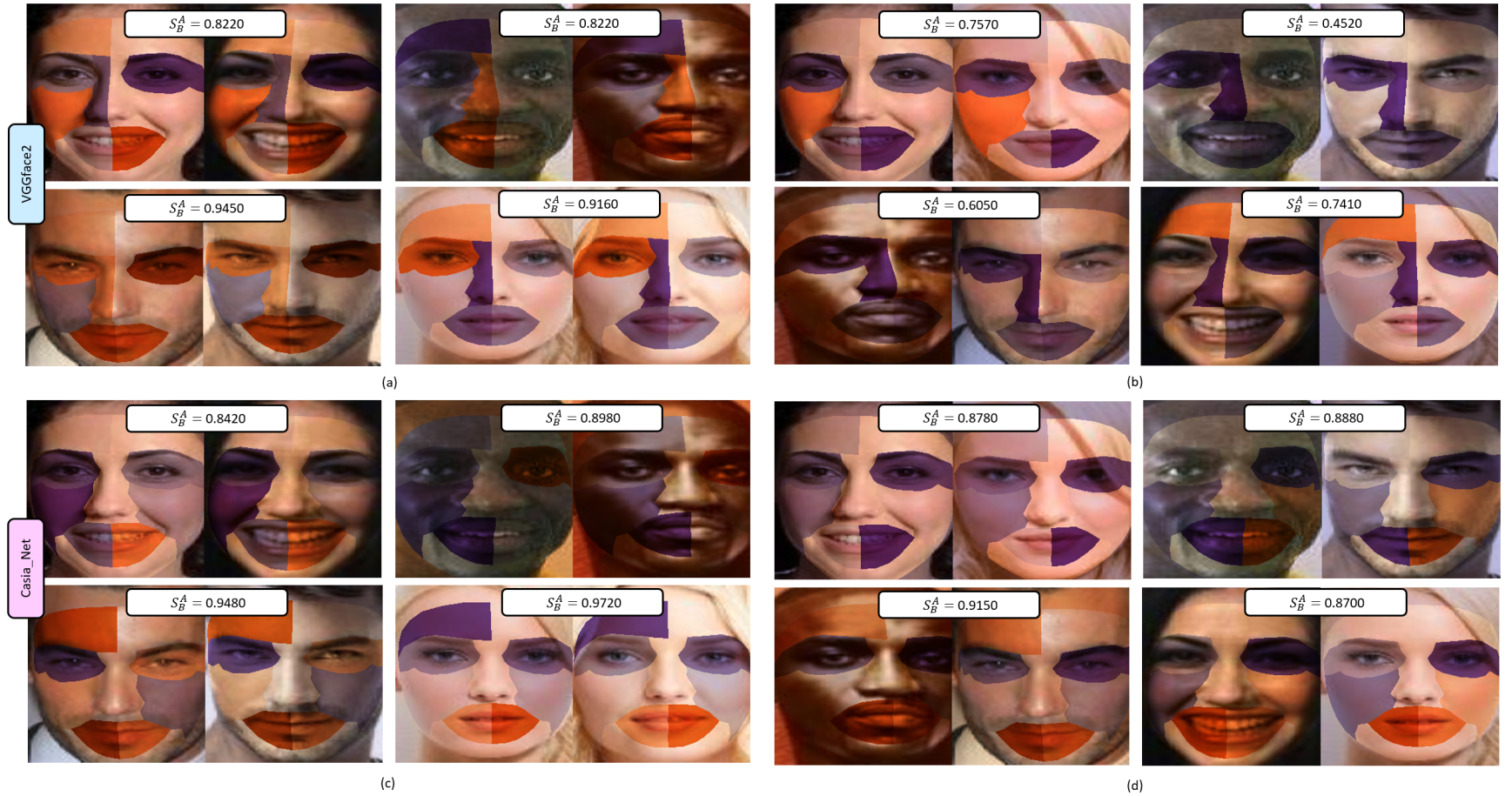}
    \caption{The S\textsubscript{AVG} map for pair examples. In sections (a) and (c), the similarity maps are generated for genuine cases, while in sections (b) and (d) for impostor ones. The examples are generated from VGGface2 (a,d) and Casia Net (b,c).}
    \label{fig: examples_results}
\end{figure}

Figure \ref{fig: maps} also features a table showcasing the sections of the face categorized as similar (orange) and dissimilar (purple), along with their respective contribution values to the final similarity map. We will focus on analyzing the mean map S\textsubscript{AVG}, which utilizes the same color scale. Regarding the nature of masking applied during perturbation, we investigated how it impacted the algorithm's output. In figure \ref{fig: masking_sensitivity}, we present two distinct case studies for both models. The examined masking types encompass black masking, random noise masking, and white masking. Upon observing the images, it becomes evident that, in general, there is minimal sensitivity to the type of masking, especially between black and white masking. The most notable deviation is associated with random noise masking, although this divergence remains relatively modest. The maps reported in this study exclusively employ black masking.
Figure \ref{fig: examples_results} presents several instances of the algorithm's output for both tested models. Specifically, sections (a) and (c) demonstrate examples where facial comparisons are made between samples of the same individual, while sections (b) and (d) involve comparisons with imposters.
Even when comparing faces of the same individual, certain areas are assessed as dissimilar, while conversely, when confronting imposters, not all areas are consistently regarded as dissimilar. The final score can offer additional insights by contextualizing which facial regions can be modified to influence the outcome.

\subsection{Experiments with Cut-and-Paste Patches}

\begin{figure}[ht]
    \centering
    \includegraphics[width=\textwidth ]{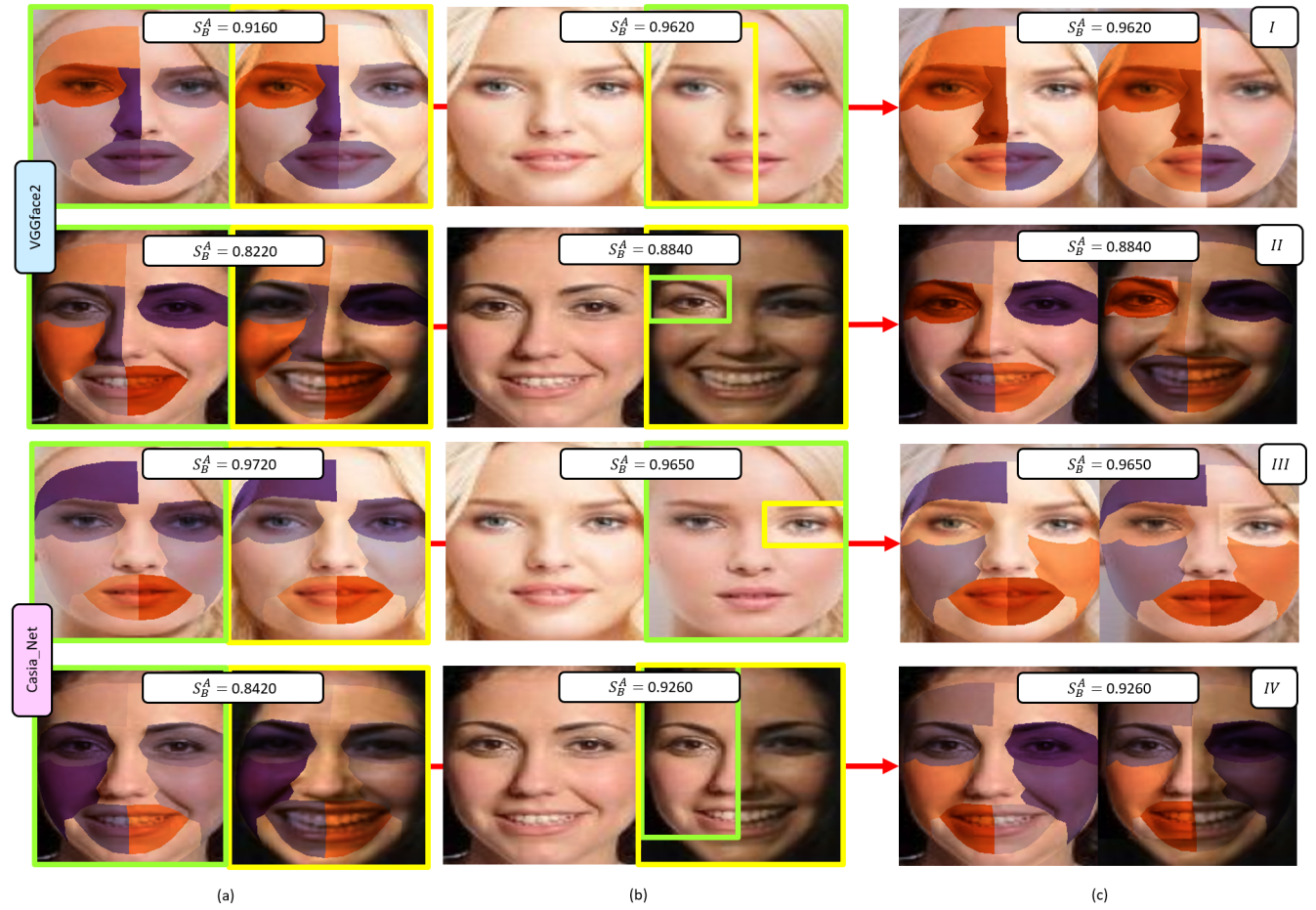}
    \caption{Cut-and-Paste patches test inspired by \cite{Jacob2021} for two samples output. In (a) the originals S\textsubscript{AVG}, in (b) the pairs after applying the Cut-and-Paste Patches test, and in (c) the new average similarity maps.}
    \label{fig: testing}
\end{figure}

We conducted a ``Cut-and-Paste Patches'' test to validate this outcome, as previously introduced by \cite{19knoche2023explainable}. This experiment assesses whether replacing specific facial regions in one image with a corresponding region from another is effectively detected by our algorithm and described with high similarity in the similarity maps. We present the results in Figure \ref{fig: testing}. Specifically, in column (a), we display the average similarity map of the two original images. In column (b), one of the two images has been altered with a patch from the other (highlighted in green-yellow). Finally, in column (c), we present the resultant output. Overall, we observe that regions previously deemed dissimilar are now perceived as similar in the modified area, accompanied by an increase in the final score.
Additionally, we notice instances where semantic areas change in their contribution despite not being included in the modification patch. The explanation for this can be that the patches do not fit the exact dimensions as the semantic areas, and in some cases, a rectangular patch, mainly centered on one point, may intersect multiple subsequently affected semantic areas. This observation underscores the sensitivity of the proposed method, particularly the segmentation carried out by Mediapipe, to facial regions. It is also noteworthy that certain areas change in color even when they have not been directly modified – for instance, the right eye in Case I, the left cheek in Case II, the left eye in Case III, and in Case IV, the patch is not entirely recognized as similar. This discrepancy can be attributed to the fact that while the test follows a part-based approach, network models tend to perceive faces holistically, implying that altering a specific patch may lead to a change in perception of the entire face and not just the modified area. This explanation aligns with the study of Jacob et al.\cite{Jacob2021}, which demonstrated that models trained on various datasets with the Thatcher effect \cite{Margaret1980} internalize a holistic perception of faces. Moreover, it is essential to note that the maps under consideration focus solely on the most influential areas, albeit their influence on the final score is limited.

\subsection{Method limitations}
\label{sec:limitation}
While Mediapipe offers valuable tools for semantically segmenting facial features, it displays a notable sensitivity to variations in facial orientation. Substantial deviations in facial pose result in increasingly dissimilar masks, leading to proportionally divergent contributions. When the masks exhibit high similarity, the simultaneous occlusion method gains coherence as it conceals identical portions of the image. Another limitation arises when comparing a profiled face with a frontal one. In such instances, Mediapipe can still identify facial features; however, the application of occlusion to both profiles loses its contextual relevance, rendering the affected areas visually less comprehensible. Consequently, the most suitable application of the method pertains to front-facing subjects with poses as closely aligned as possible.

\section{Conclusion and Future directions}
\label{sec:conclusion}
In this paper, we have initiated an effort to bridge the gap between computer and human vision, with the primary goal of improving the interpretability of facial verification algorithms. We sought to gain insight into how machines perceive the semantic aspects of human faces during verification, ultimately aligning the system's output score more closely with human reasoning.

We employed the Mediapipe tool to identify distinct semantic regions on the human face to achieve this. These regions, representing human-conceptual knowledge, provided a comprehensive view of the critical concepts for our models. Leveraging this knowledge, we selected a subset of the most significant semantic areas for the models. We also introduced a perturbation algorithm that generated similarity maps, revealing how the models under examination perceived these concepts as either \textit{similar} or \textit{dissimilar}.

By contextualizing the system's output score, we can align it more closely with human reasoning.
However, it is essential to note that our work is currently limited to experimentation. As a result, future research directions could include exploring different segmentation methods, conducting experiments across diverse models, comparing various methods to ours, or adapting them to our approach. Additionally, including a user evaluation component could further validate and enhance the effectiveness of our work.

\bibliographystyle{elsarticle-num}

\bibliography{main}

\end{document}